\documentclass[runningheads]{llncs}

\usepackage{eccv}

\usepackage{eccvabbrv}

\usepackage{graphicx}
\usepackage{booktabs}
\usepackage{amsmath}
\usepackage{amsfonts}
\usepackage{multirow}
\usepackage{stfloats}
\usepackage{siunitx}
\usepackage{collcell}
\usepackage{xcolor}

\usepackage[accsupp]{axessibility}  

\newcommand{\best}[1]{\textcolor{red}{\textbf{#1}}}
\newcommand{\second}[1]{\textcolor{blue}{\underline{#1}}}

\usepackage{hyperref}

\usepackage{orcidlink}

\begin{document}

\title{HLGFA: High–Low Resolution Guided Feature Alignment for Unsupervised Anomaly Detection}

\titlerunning{HLGFA: High-Low Resolution Guided Feature Alignment}

\author{
Han Zhou\inst{1} \and
Yuxuan Gao\inst{1} \and
Yinchao Du\inst{1} \and
Xuezhe Zheng\inst{1}
}

\authorrunning{H. Zhou et al.}

\institute{
Innolight Technology Research Institute\\
\email{\{han.zhou, yuxuan.gao, yinchao.du, xuezhe.zheng\}@innolight.com}
}

\maketitle

\begin{abstract}
Unsupervised industrial anomaly detection (UAD) is critical for automated visual inspection in modern manufacturing, where defect samples are scarce and annotation is costly. Existing reconstruction- and memory-based approaches, while promising, often suffer from unstable anomaly responses and high inference complexity. We propose HLGFA, a high--low resolution guided feature alignment framework that learns normality through cross-resolution feature consistency, without explicit reconstruction or external memory. A shared frozen backbone extracts multi-level features from dual-resolution inputs; high-resolution features are decomposed into structure and detail priors, which guide low-resolution feature refinement via conditional modulation and gated residual correction. At inference, anomalies naturally emerge as regions where cross-resolution alignment breaks down. A noise-aware augmentation strategy is further introduced to suppress nuisance-induced false positives in real industrial environments. Experiments on MVTec AD and VisA demonstrate the effectiveness of HLGFA, achieving 98.0\% pixel-level AUROC on MVTec AD, and 97.9\% pixel-level AUROC on VisA, delivering consistent detection and localization performance across diverse industrial defect categories.

\end{abstract}

\section{Introduction}
\label{sec:intro}

Industrial anomaly detection (IAD) is a fundamental component of modern manufacturing quality control, aiming to automatically identify defects or irregularities without human intervention. With the rapid deployment of automated production lines, reliable visual inspection systems have become essential across a wide range of industries, where manual inspection is often inefficient, error-prone, and difficult to scale.

Despite substantial progress in computer vision, IAD remains challenging due to several inherent factors: the extreme imbalance between normal and defective samples, the large diversity and unpredictability of defect appearances, and the presence of complex industrial backgrounds that can obscure or mimic defects. Moreover, real-world applications demand high precision and recall, as missed or false detections may lead to significant economic loss or safety risks.

Unsupervised anomaly detection (UAD) has therefore attracted increasing attention, as it learns normal patterns solely from defect-free samples and identifies anomalies as deviations during inference. Existing UAD approaches can be broadly grouped into reconstruction-based methods~\cite{ZAVRTANIK2021107706}, feature-based methods such as PatchCore~\cite{roth2022totalrecallindustrialanomaly} and UniAD~\cite{you2022unifiedmodelmulticlassanomaly}, and more recent zero-shot or few-shot paradigms leveraging vision-language models~\cite{jeong2023winclipzerofewshotanomalyclassification,jin2025dualinterrelateddiffusionmodelfewshot,gao2025metauasuniversalanomalysegmentation}. While effective, many existing methods struggle to simultaneously preserve global structural consistency and local detail fidelity, especially when defects vary significantly in scale.

\begin{figure}
    \centering
    \includegraphics[width=0.98\textwidth]{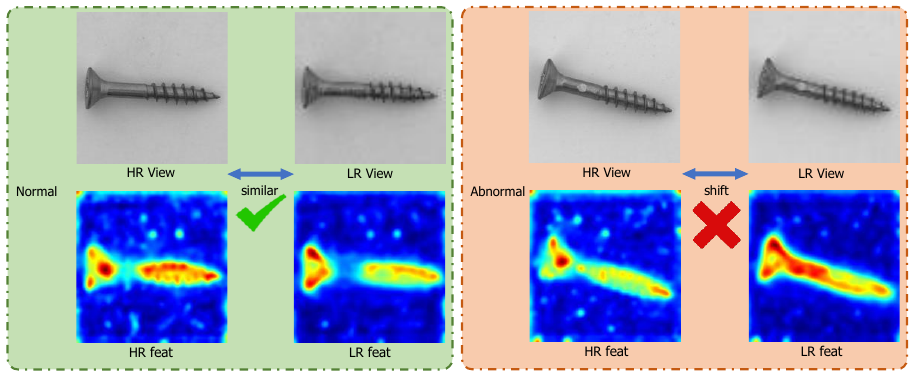}
    \caption{
    Visualization of feature responses extracted by a pretrained backbone under different resolutions. Normal samples show consistent activation patterns across high- and low-resolution views, while anomalous samples exhibit pronounced response shifts after resolution reduction due to the degradation of fine-grained structural cues.
    }
    \label{fig:motivation}
\end{figure}

A key observation in industrial inspection is that normal patterns exhibit stable feature responses under resolution variation, whereas anomalous regions are considerably more sensitive to resolution reduction. When images are downsampled, the global structure of normal objects is largely preserved, while defect-related cues—typically local and irregular—tend to degrade or shift. This suggests that cross-resolution feature consistency provides an intrinsic cue for normality, whereas pronounced inconsistencies naturally indicate anomalies. As illustrated in Fig.~\ref{fig:motivation}, normal samples maintain consistent responses across resolutions, while anomalous samples exhibit clear response shifts, offering a principled signal for unsupervised anomaly detection without explicit defect modeling.

Motivated by this observation, we propose HLGFA, a high--low resolution guided feature alignment framework for unsupervised industrial anomaly detection. Instead of relying on pixel-level reconstruction, HLGFA models normality by enforcing cross-resolution feature consistency between high-resolution and low-resolution views of normal samples.

Given an input image, dual-resolution features are extracted by a shared frozen backbone and aligned through a learnable guidance module that exploits the asymmetric characteristics of high- and low-resolution representations. High-resolution features are further decomposed into structure and detail priors, which jointly guide the refinement of low-resolution features via conditional modulation and gated residual correction. During inference, anomalies are detected as regions where cross-resolution alignment breaks down. In addition, a noise-aware data augmentation strategy is introduced to improve robustness against nuisance patterns commonly observed in industrial environments.

The main contributions of this work are summarized as follows:
\begin{itemize}
\item We propose an unsupervised anomaly detection framework that identifies anomalies through cross-resolution feature inconsistency.
\item We design a structure--detail decoupled guided alignment module that enables stable cross-resolution feature alignment without updating backbone parameters.
\item We introduce a noise-aware data augmentation strategy to enhance robustness and reduce false positives in industrial scenarios.
\end{itemize}

The remainder of this paper is organized as follows. Section~\ref{sec:related} reviews related work, Section~\ref{sec:method} presents the proposed method, Section~\ref{sec:experiments} reports experimental results, and Section~\ref{sec:conclusion} concludes the paper.

\section{Related Works}
\label{sec:related}

Advancements in industrial anomaly detection have been driven by diverse methodologies focused on identifying defects with minimal labeled data. This section reviews key works in the field, which fall into the following categories.

\subsection{Reconstruction-based Anomaly Detection}
Reconstruction-based methods identify anomalies by learning to model the distribution of normal data and defect deviations through reconstruction errors. Early generative works, such as SPADE \cite{park2019semanticimagesynthesisspatiallyadaptive}, introduced spatially-adaptive normalization for high-fidelity semantic image synthesis, influencing subsequent methods that leverage conditional generation for data augmentation or reconstruction. A notable limitation of standard auto-encoders is their tendency to generalize too well, potentially reconstructing anomalies. To address this, Dong Gong et al.\cite{gong2019memorizingnormalitydetectanomaly} proposed a memory-augmented deep autoencoder (MemAE), which constrains the latent representation by correlating it with a memory bank of prototypical normal patterns, effectively preventing the reconstruction of anomalous regions.

Building on the principle of constrained reconstruction, \cite{ZAVRTANIK2021107706} proposed RIAD (Reconstruction by Inpainting for Visual Anomaly Detection), a novel approach that frames anomaly detection as a self-supervised reconstruction-by-inpainting problem. Unlike standard auto-encoding methods, RIAD randomly removes and reconstructs partial image regions from partial inpainting. By forcing the model to learn from incomplete data, RIAD achieves state-of-the-art results on challenging anomaly detection benchmarks.

Further advancing reconstruction-based paradigms, \cite{jin2025dualinterrelateddiffusionmodelfewshot} introduced the Dual-Interrelated Diffusion Model for few-shot anomaly image generation. Their approach tackles the critical challenge of data scarcity in industrial defect detection by using dual-interrelated diffusion processes to generate anomaly images with limited examples. This enables more effective anomaly detection, as the model learns from very few anomaly samples through sophisticated diffusion modeling techniques.

\subsection{Feature-based Anomaly Detection}
Feature-based methods leverage pre-trained representations and feature matching strategies for anomaly detection. A foundational work in this area is PaDiM (Patch Distribution Modeling)\cite{defard2020padimpatchdistributionmodeling}. PaDiM utilizes a pre-trained convolutional neural network to extract patch-level features, models their multivariate Gaussian distributions from normal training images, and detects anomalies by computing the Mahalanobis distance of test patches to the learned normal distribution, achieving robust localization performance. Extending this paradigm, PatchCore \cite{roth2022totalrecallindustrialanomaly} focuses on achieving total recall in industrial anomaly detection by constructing a memory bank of nominal patch features and using a nearest-neighbor search for scoring. This approach maximizes detection sensitivity while maintaining precision, addressing the practical requirement of minimizing false negatives in industrial inspection scenarios.

Further advancing this paradigm, UniAD~\cite{you2022unifiedmodelmulticlassanomaly} proposes a unified framework for multi-class anomaly detection via memory bank and prototype learning, achieving robust discrimination across diverse object categories. In contrast, RD4AD~\cite{deng2022anomalydetectionreversedistillation} adopts a reverse distillation strategy, where a student network is trained to reconstruct teacher features from a compact one-class bottleneck. Since the student is optimized solely on normal patterns, anomalous regions naturally manifest as high reconstruction discrepancies at inference.

\subsection{Zero-shot and Few-shot Anomaly Detection}
Recent studies have explored leveraging large-scale vision--language models to perform anomaly detection with little or no task-specific training data. In particular, the emergence of CLIP~\cite{radford2021learningtransferablevisualmodels} has inspired a line of work that formulates anomaly detection as a vision--language similarity or alignment problem.

Early efforts such as WinCLIP~\cite{jeong2023winclipzerofewshotanomalyclassification} demonstrate that carefully designed prompts enable effective zero-shot and few-shot anomaly recognition. Building on this idea, AnomalyCLIP~\cite{zhou2023anomalyclip} introduces anomaly-aware prompt learning to improve pixel-level localization, while CLIP-AD~\cite{chen2024clipadlanguageguidedstageddualpath} contrasts normal and abnormal semantic embeddings to avoid category-specific training. PromptAD~\cite{li2024promptadlearningpromptsnormal} further explores prompt-conditioned feature alignment under few-shot supervision for anomaly segmentation.

Beyond prompt-based formulations, several methods investigate more structured representations to enhance generalization. April-GAN~\cite{chen2023aprilganzerofewshotanomalyclassification} synthesizes diverse anomaly patterns via adversarial generation, whereas MetaUAS~\cite{gao2025metauasuniversalanomalysegmentation} adopts a meta-learning paradigm to enable rapid adaptation across categories. MAEDAY~\cite{schwartz2024maedaymaezeroshot} combines masked autoencoding with CLIP to learn robust normal representations without explicit anomaly supervision. Overall, these approaches highlight the promise of vision--language models in reducing annotation requirements while achieving competitive anomaly detection performance.

\section{Method}
\label{sec:method}

To overcome the key challenges of modeling anomalies across multiple scales, the limited sensitivity to fine-grained defects in reconstruction-based methods, and the overfitting of normal patterns in unsupervised industrial anomaly detection, this paper proposes an unsupervised anomaly detection framework based on cross-resolution feature alignment with guided representation learning. By exploiting the asymmetric representational characteristics of high- and low-resolution features, this approach explicitly converts cross-resolution inconsistency into a reliable anomaly signal for detection.

\begin{figure}
    \centering
    \includegraphics[width=0.98\textwidth]{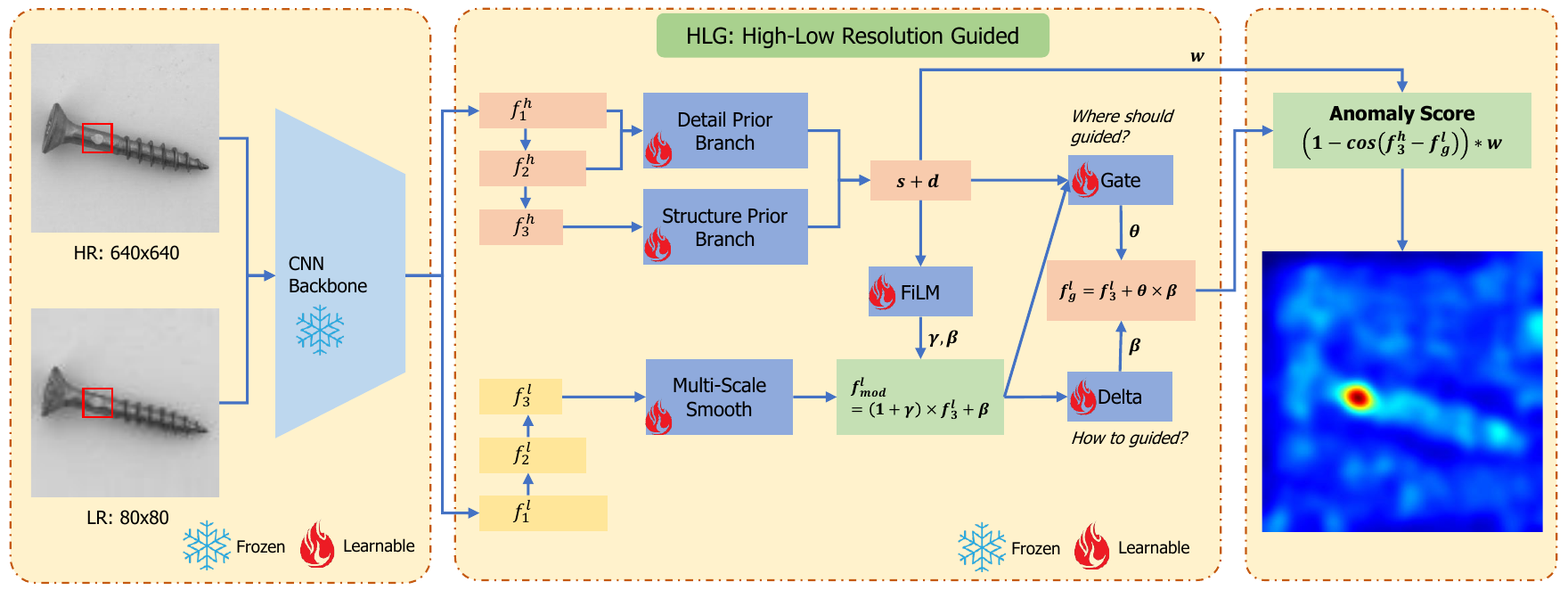}
    \caption{
    High-resolution (HR) and low-resolution (LR) images are processed by a shared frozen backbone to extract multi-scale features.The learnable HLGFA module performs structure-guided refinement of low-resolution features using high-resolution representations.Anomalies are detected as regions where cross-resolution feature alignment fails.
    }
    \label{fig:hlg_framework}
\end{figure}

\subsection{High--Low Resolution Feature Guide}

As illustrated in Fig.~\ref{fig:hlg_framework}, the proposed framework adopts a dual-resolution feature guidance strategy to leverage the asymmetric representational characteristics of high- and low-resolution (HR and LR) features. High-resolution (HR) representations provide structural and detail priors to guide the refinement of low-resolution (LR) features. This is because HR features preserve richer structural and spatial information, while LR features offer a more compact and noise-tolerant abstraction.

Given an input image $x$, we construct a high-resolution view $x^{h}$ and a low-resolution view $x^{l}$ via down-sampling. Both views are processed by a shared backbone encoder $B(\cdot)$, producing multi-stage feature representations $\{f^{h}_s\}$ and $\{f^{l}_s\}$ at different semantic levels.

Due to the inherent resolution mismatch, LR features are first spatially resized to match their HR counterparts:
\begin{equation}
\bar{f}^{l}_s = \mathcal{U}_s(f^{l}_s),
\end{equation}
where $\mathcal{U}_s(\cdot)$ denotes a stage-specific up-sampling operator. This operation ensures spatial correspondence while preserving the compact semantics of LR features.

Rather than directly enforcing feature similarity, which is often unstable under resolution discrepancies, we introduce a learnable guided alignment operator $\mathcal{G}_s(\cdot)$. This operator refines LR features by injecting guidance from HR representations, providing a more effective and adaptive alignment:
\begin{equation}
\hat{f}^{l}_s = \mathcal{G}_s(\bar{f}^{l}_s, f^{h}_s).
\end{equation}
In practice, HR features serve as guidance signals modulating and correcting LR features. Mean while LR features remain the primary carriers of anomaly-sensitive representations.

During inference time, anomalies are detected as spatial regions where cross-resolution consistency is violated. 
The anomaly score is computed based on the cosine dissimilarity between the reconstructed low-resolution feature and the corresponding high-resolution feature:
\begin{equation}
\mathcal{A}(x) =\sum_{s}\left(1 - \frac{\left\langle \hat{f}^{l}_s(x), f^{h}_s(x) \right\rangle}{\left\| \hat{f}^{l}_s(x) \right\|_2\left\| f^{h}_s(x) \right\|_2}\right),
\end{equation}
where larger cosine dissimilarities indicate stronger abnormality. This formulation directly transforms cross-resolution feature inconsistency into a reliable anomaly signal, as visualized in Fig.~\ref{fig:hlg_framework}.

\subsection{Structure--Detail Decoupled Guidance}

\begin{figure}
    \centering
    \includegraphics[width=0.98\textwidth]{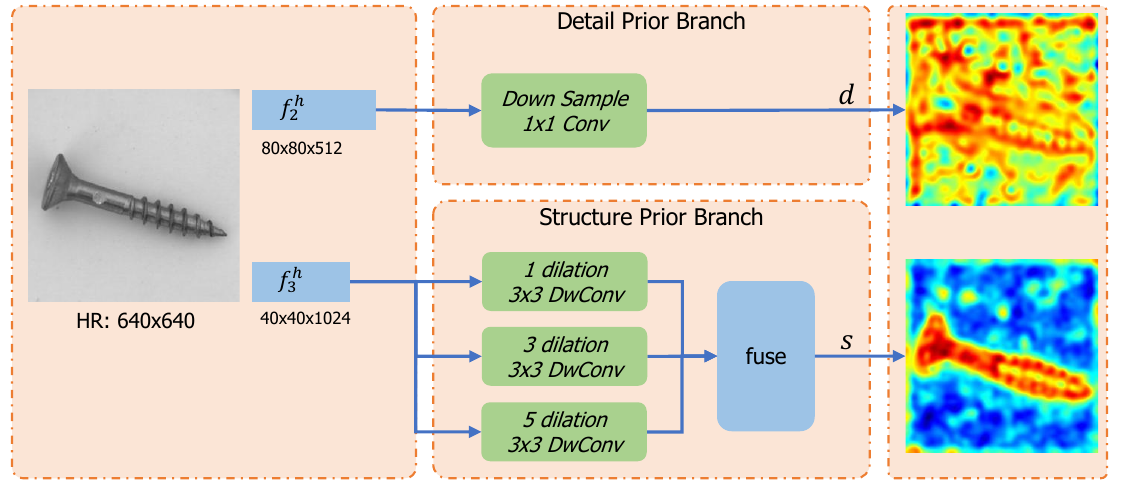}
    \caption{
    Illustration of the proposed structure--detail decoupled guidance. High-resolution (HR) features are decomposed into a structure prior and a detail prior. The structure prior captures stable semantic layouts via multi-scale depthwise convolutions, while the detail prior preserves informative local cues through lightweight spatial alignment and channel projection, enabling stable cross-resolution guidance.
    }
    \label{fig:hlg_framework_detail}
\end{figure}

A key challenge in cross-resolution feature alignment lies in the entangled nature of high-resolution (HR) representations, where global structure and fine-grained details coexist in the same feature space. Directly using HR features as guidance often leads to unstable supervision, as local noise or repetitive textures may dominate the alignment signal.

To mitigate this issue, we explicitly decompose HR features into two complementary priors: a \emph{structure prior} and a \emph{detail prior}, as illustrated in Fig.~\ref{fig:hlg_framework_detail}. The structure prior is extracted from deeper HR features using multi-scale depthwise convolutions to model stable global layouts, while the detail prior is derived from shallower HR features through spatial alignment and lightweight channel projection, retaining informative local cues while suppressing excessive high-frequency noise. The two priors are combined into a unified guidance representation:
\begin{equation}
g_s = s_s + d_s,
\end{equation}
where $s_s$ and $d_s$ denote the structure and detail priors, respectively.

As shown in Fig.~\ref{fig:hlg_framework}, low-resolution (LR) features are first spatially aligned and stabilized, and then modulated via a FiLM-based transformation conditioned on the structure--detail guidance:
\begin{equation}
\tilde{f}^{l}_s = \mathrm{FiLM}(f^{l}_s \mid g_s),
\end{equation}
where $\mathrm{FiLM}(\cdot)$ denotes a feature-wise affine modulation conditioned on $g_s$.

Rather than enforcing strict feature-level consistency, we adopt a lightweight gated residual correction mechanism. The refined LR features are obtained as:
\begin{equation}
\hat{f}^{l}_s = f^{l}_s + \mathcal{G}(g_s, \tilde{f}^{l}_s),
\end{equation}
where $\mathcal{G}(\cdot)$ denotes a shallow gated residual predictor that adaptively determines \emph{where} and \emph{how} HR guidance should be injected. This formulation avoids overly restrictive pixel-wise constraints and enables flexible, data-driven cross-resolution alignment.

\begin{figure}
    \centering
    \includegraphics[width=0.98\textwidth]{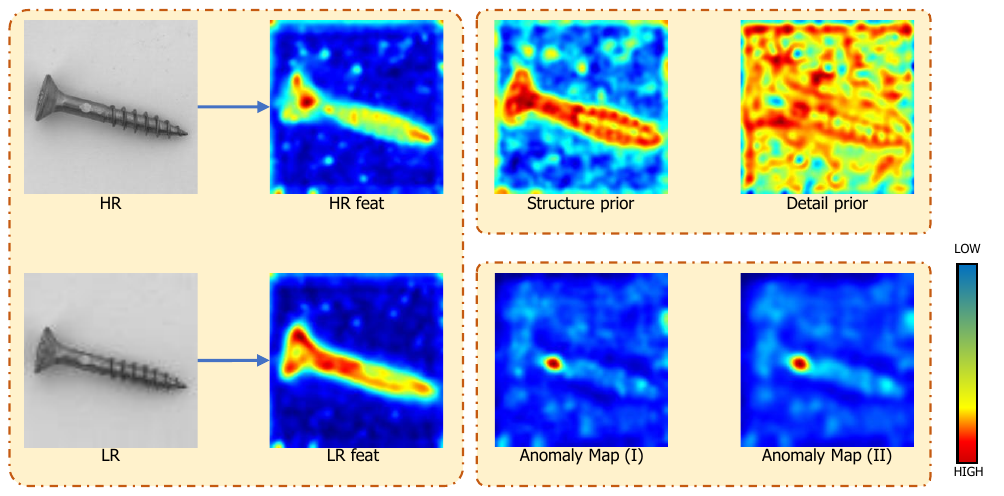}
    \caption{
    Visualization of the proposed structure--detail decoupled guidance and structure-based reliability modulation. HR and LR images are encoded into multi-scale features. During inference, anomaly maps derived from cross-resolution discrepancies are further modulated by a structure-based reliability weight, which suppresses spurious responses in structurally unstable regions.
    }
    \label{fig:lr_g}
\end{figure}

While structure--detail guidance improves cross-resolution alignment during training, we further exploit the stability of the structure prior to estimate spatial reliability during inference, as illustrated in Fig.~\ref{fig:lr_g}. Intuitively, anomaly responses caused by unreliable or noisy guidance should be suppressed, even if cross-resolution discrepancies are observed.

Specifically, we compute a local structural consistency score based on the normalized guidance representation:
\begin{equation}
\mathrm{sim}(x) =
\mathbb{E}_{p \in \mathcal{N}(x)}
\left[
\left\langle
\tilde{g}_s(x), \tilde{g}_s(p)
\right\rangle
\right],
\end{equation}
where $\mathcal{N}(x)$ denotes the local neighborhood of $x$ and $\tilde{g}_s(x)=g_s(x)/\|g_s(x)\|_2$.

The final reliability-aware anomaly response is obtained by:
\begin{equation}
\mathcal{A}'(x) =
\sigma\!\left(
\frac{\mathrm{sim}(x) - \tau}{\delta}
\right)
\cdot \mathcal{A}(x),
\end{equation}
where $\tau$ and $\delta$ control the consistency threshold and transition smoothness, respectively.

\subsection{Noise-Aware Data Augmentation}

\begin{figure}
    \centering
    \includegraphics[width=0.98\textwidth]{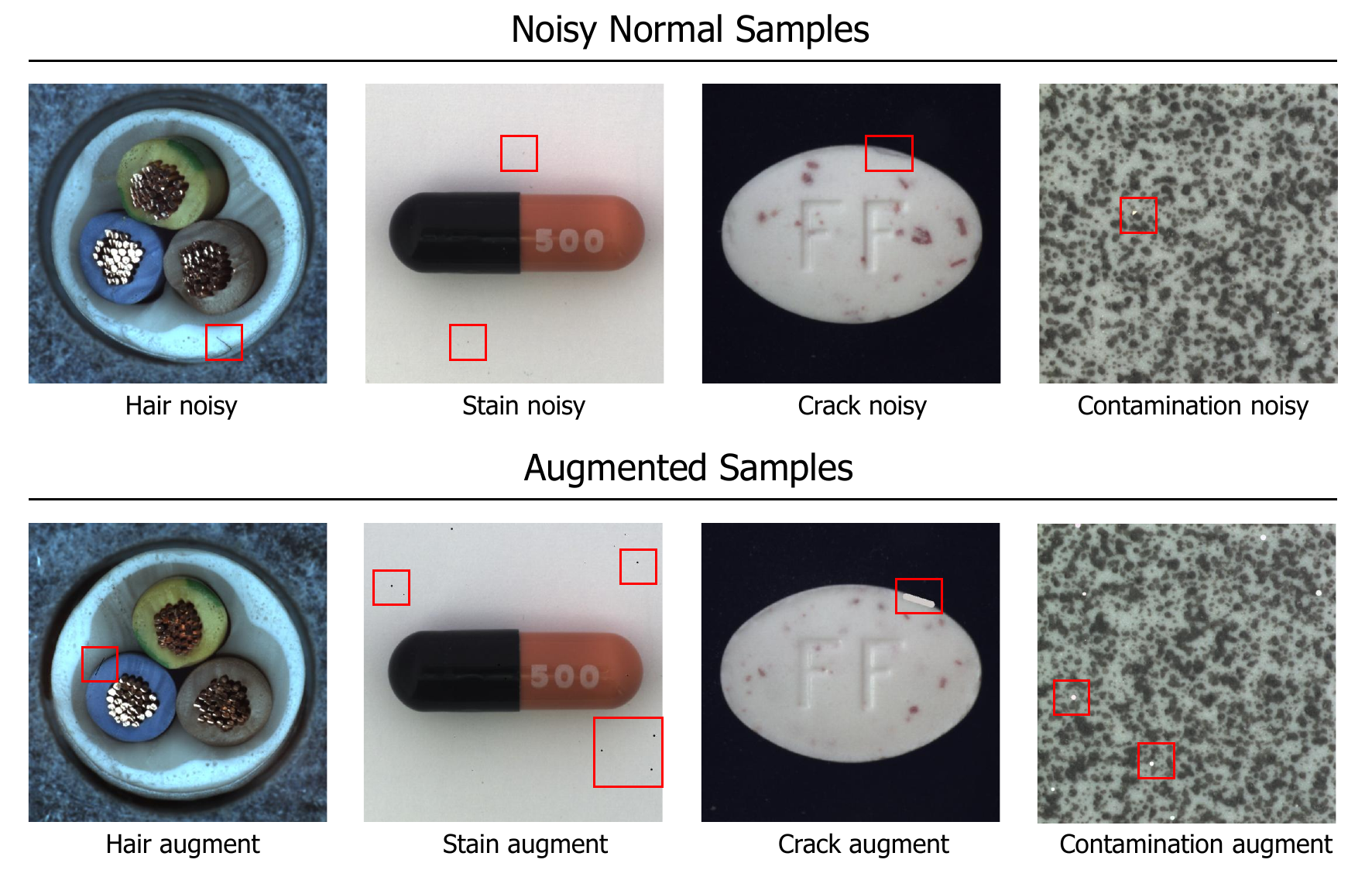}
    \caption{
    The top row shows typical nuisance patterns commonly observed in defect-free products, including hairs, stains, cracks, and contamination noise.
    The bottom row illustrates our noise-aware augmentation strategy, where sparse point noise and structured stripe noise are synthetically injected into normal samples to simulate real-world contamination.
    }
    \label{fig:data_aug}
\end{figure}

In real-world industrial inspection, normal samples often contain unavoidable nuisance factors such as hairs, stains, or contamination noise. Although visually irrelevant, these patterns can interfere with cross-resolution feature alignment, leading to false positives. As shown in Fig.~\ref{fig:data_aug}, even defect-free products may exhibit minor surface irregularities that produce strong local activations, causing the model to incorrectly interpret them as anomalies.

To improve robustness, we apply noise-aware data augmentation during training by injecting sparse point noise and structured stripe noise into normal samples, mimicking common contamination patterns. Noise is applied consistently to both HR and LR views, encouraging the model to focus on stable structural semantics rather than local noise, thereby reducing false positive detections.

\subsection{Backbone Selection}

We empirically observe that ResNet-style backbones, especially Wide-ResNet~\cite{zagoruyko2016wideresnet} variants, consistently outperform DenseNet~\cite{huang2018denselyconnectedconvolutionalnetworks} and ConvNeXt~\cite{liu2022convnet2020s} architectures within the proposed framework. This is likely due to the spatially diffused feature propagation enabled by residual connections, which promotes smooth and coherent feature responses across layers.

Such properties are particularly beneficial for cross-resolution alignment, where reliable guidance depends on stable and spatially coherent representations. In contrast, architectures that emphasize aggressive feature reuse or localized depthwise convolutions tend to amplify texture-level noise, which is detrimental to robust anomaly detection, leading to less reliable results.

\subsection{Training Objective}

All training objectives are computed exclusively on normal samples to ensure the model learns the characteristics of normal data. The core objective enforces cross-resolution feature alignment by minimizing a cosine similarity loss between the high-resolution features $f^{h}$ and the aligned low-resolution features $\hat{f}^{l}$:
\begin{equation}
\mathcal{L}_{\text{align}} =
\mathbb{E}\!\left[
1 - \frac{\langle f^{h}, \hat{f}^{l} \rangle}
{\|f^{h}\|_2 \, \|\hat{f}^{l}\|_2}
\right].
\end{equation}
This formulation encourages resolution-invariant representations while remaining insensitive to feature magnitude, making it particularly suitable when using a frozen backbone.

To improve robustness under noisy or ambiguous conditions, we further augment the alignment objective with several lightweight auxiliary regularizers. Specifically, we introduce (i) a focal-weighted $\ell_1$ feature consistency term to emphasize misaligned regions, (ii) a distribution-level consistency constraint based on Jensen--Shannon divergence by treating channel-wise normalized features as probability distributions, and (iii) a Gram-matrix matching loss to encourage second-order structural consistency between high- and low-resolution features. In addition, a classification-based auxiliary loss is optionally employed to provide weak semantic regularization.

The overall training objective is defined as:
\begin{equation}
\mathcal{L} = \mathcal{L}_{\text{align}} +
\lambda_{\ell_1}\mathcal{L}_{\ell_1} +
\lambda_{\text{JS}}\mathcal{L}_{\text{JS}} +
\lambda_{\text{Gram}}\mathcal{L}_{\text{Gram}} +
\lambda_{\text{cls}}\mathcal{L}_{\text{cls}},
\end{equation}
where the auxiliary terms mainly serve as stabilizers during training. Unless otherwise specified, $\mathcal{L}_{\text{align}}$ is always enabled, while the remaining terms are assigned modest weights and activated only when necessary.

\section{Experiments}
\label{sec:experiments}

\begin{figure}
    \centering
    \includegraphics[width=0.98\textwidth]{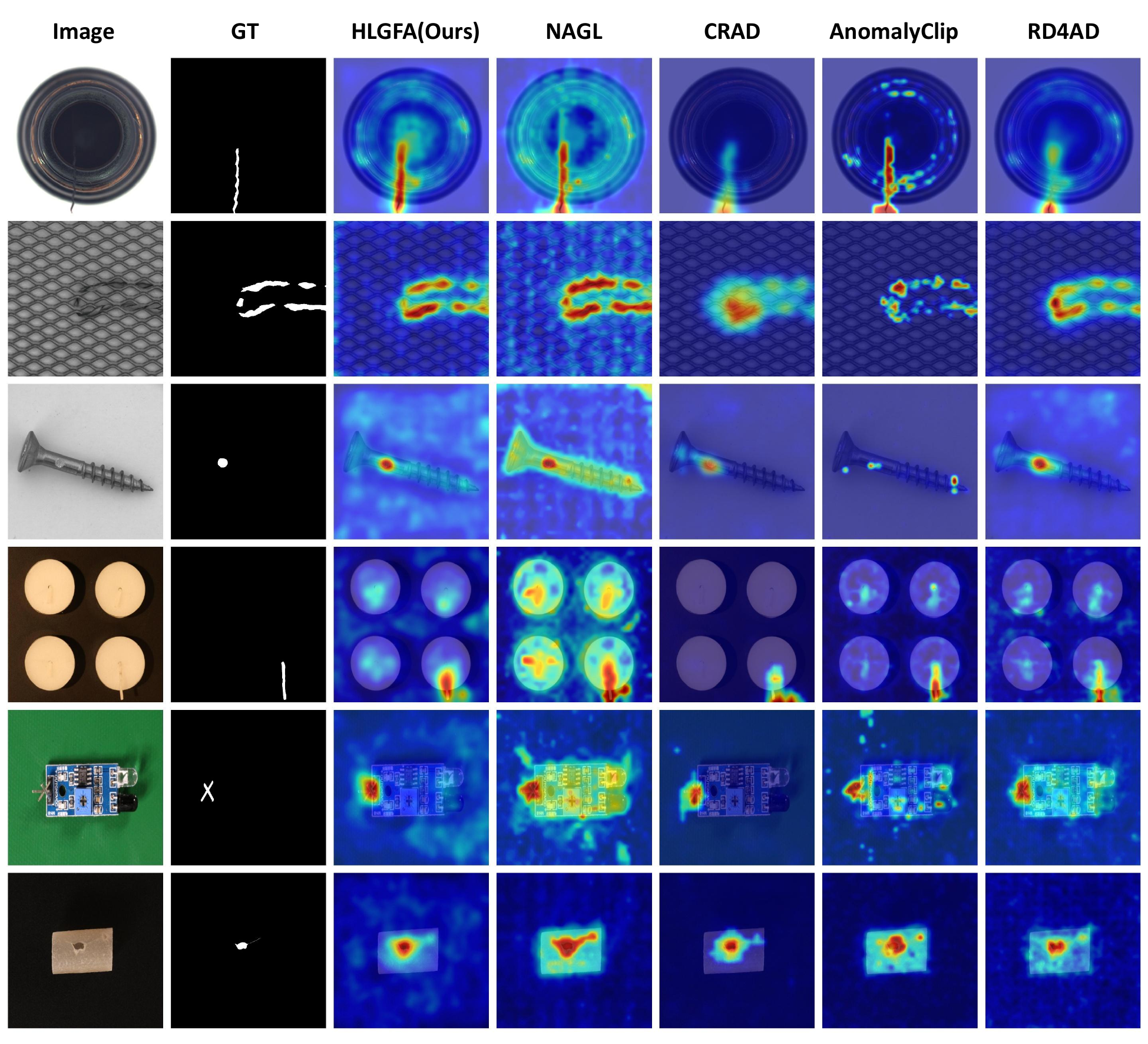}
    \caption{
    Qualitative comparison of anomaly localization results on the MVTec AD and VisA dataset. From left to right: input image, ground-truth mask (GT), HLGFA (ours), NGAL, CRAD, AnomalyCLIP, and RD4AD. HLGFA produces more compact and accurate anomaly responses that align better with the ground-truth regions, while suppressing spurious activations on normal areas.
    }
    \label{fig:exp}
\end{figure}

\subsection{Experimental Setup}

We evaluate on \textbf{MVTec AD}~\cite{mvtec_ad} and \textbf{VisA}~\cite{zou2022spotthedifferenceselfsupervisedpretraininganomaly}. Following the unsupervised protocol, only normal images are used for training in a category-agnostic manner. We report \textbf{pixel-level} metrics (AUC-P, AP-P, F1-P, PRO) and \textbf{image-level} metrics (AUC-I, AP-I, F1-I) following the official evaluation protocol.

We use a \textbf{frozen} Wide-ResNet-50 backbone pretrained on ImageNet with $640 \times 640$ input resolution. Training uses Adam optimizer (learning rate $1\times10^{-3}$ to $1\times10^{-4}$ with cosine annealing) for 100 epochs with batch size 32 on normal samples only. Inference requires no memory bank or reference samples---anomaly maps are computed directly from cross-resolution feature inconsistency.

\subsection{Overall Comparison with State-of-the-Art Methods}

Table~\ref{tab:comparison} presents quantitative comparisons on both MVTec AD and VisA benchmarks against representative state-of-the-art approaches, including RD4AD\cite{deng2022anomalydetectionreversedistillation}, AnomalyCLIP\cite{zhou2023anomalyclip}, CRAD\cite{lee2024continuousmemoryrepresentationanomaly}, and NAGL\cite{wang2025normalabnormalguidedgeneralistanomaly}.

As shown in Table~\ref{tab:comparison}, HLGFA achieves competitive performance across both datasets and demonstrates clear advantages in anomaly localization. On MVTec AD, our method achieves 97.4\% ROC-I and 98.0\% ROC-P, while obtaining the best pixel-level AP of 65.3\%, significantly outperforming existing methods. This improvement indicates that HLGFA provides more accurate pixel-wise anomaly localization and produces more precise segmentation maps.

\begin{table}
\centering
\caption{
Quantitative comparison on MVTec AD and VisA. We report image-level ROC-AUC (ROC-I), pixel-level ROC-AUC (ROC-P), and pixel-level AP (AP-P). \best{Red} denotes the best result and \second{blue} denotes the second-best result.
}
\label{tab:comparison}
\setlength{\tabcolsep}{6pt}
\renewcommand{\arraystretch}{1.15}
\resizebox{\textwidth}{!}{
\begin{tabular}{lccccc ccccc}
\toprule
\multirow{2}{*}{Method}
& \multicolumn{5}{c}{\textbf{MVTec AD}}
& \multicolumn{5}{c}{\textbf{VisA}} \\
\cmidrule(lr){2-6}
\cmidrule(lr){7-11}
& ROC-I & AP-I & ROC-P & AP-P & PRO-P
& ROC-I & AP-I & ROC-P & AP-P & PRO-P \\
\midrule
RD4AD (CVPR'22)
& 92.1 & 96.2 & 96.3 & 50.1 & 90.9
& \second{93.5} & \second{91.2} & 69.9 & 2.6 & 46.9 \\
AnomalyCLIP (ICLR'24)
& 91.6 & 96.4 & 91.1 & 34.5 & 81.4
& 82.0 & 85.3 & 95.3 & 22.2 & \second{86.7} \\
CRAD (ECCV'24)
& \best{98.2}& \second{98.9} & \second{97.8} & 52.6 & \second{91.0}
& \best{95.1} & \best{95.6} & \best{98.1} & 33.7 & 81.0 \\
NGAL (NeurIPS'25)
& 94.5 & 97.1 & 96.1 & \second{58.8} & \best{91.5}
& 87.1 & 88.6 & 97.3 & \second{38.9} & \best{91.5} \\
\midrule
HLGFA (Ours)
& \second{97.4} & \best{99.1} & \best{98.0} & \best{65.3} & 90.2
& 88.6 & 90.0 & \second{97.9} & \best{40.7} & 83.8 \\
\bottomrule
\end{tabular}
}
\end{table}

On the more challenging VisA benchmark, HLGFA achieves 88.2\% ROC-I and 97.8\% ROC-P, while obtaining the highest pixel-level AP of 40.6\% among all compared methods. The consistent advantage in pixel-level AP across both benchmarks highlights the robustness of HLGFA in suppressing false positives and capturing fine-grained defect regions.

These quantitative improvements are further supported by the qualitative results shown in Fig.~\ref{fig:exp}. Compared with prior methods, HLGFA produces more compact and accurate anomaly maps that better align with ground-truth defect regions while avoiding spurious responses on normal structures.

Notably, our approach achieves competitive or superior localization accuracy without relying on reconstruction, diffusion sampling, or memory banks during inference.

\subsection{Ablation Study}

\subsubsection{Feature Alignment Loss}
To investigate the contribution of each loss term, we conduct ablation experiments on MVTec AD by progressively enabling individual components of the feature alignment objective. All variants share identical training settings with Wide-ResNet50~\cite{zagoruyko2016wideresnet} as the frozen backbone, differing only in loss composition. As shown in Table~\ref{tab:ablation_loss}, cosine similarity alone already provides a competitive baseline. Among the auxiliary terms, the $\ell_1$ loss yields the most notable improvement, increasing AP-P by 3.9 points, which suggests that explicit element-wise consistency helps stabilize cross-resolution correspondence.

The Gram matrix loss further improves pixel-level localization by modeling higher-order feature correlations. In contrast, Jensen--Shannon (JS) divergence alone slightly degrades the image-level metric, implying that strong global distribution constraints may suppress subtle anomaly signals. When combined with other losses, however, JS divergence introduces mild distributional smoothing that benefits overall performance. The complete objective achieves the best results across all metrics, indicating that these regularization terms play complementary roles in aligning cross-resolution representations.

\begin{table}
\centering
\caption{Ablation study on the composition of the feature alignment loss on MVTec AD. All models use Wide-ResNet50 as the backbone.}
\label{tab:ablation_loss}
\setlength{\tabcolsep}{6pt}
\resizebox{\textwidth}{!}{%
\begin{tabular}{c c c c c c c c c}
\toprule
Cos & JS & Gram & L1 
& AUROC-I $\uparrow$ & AP-I $\uparrow$ 
& AUROC-P $\uparrow$ & AP-P $\uparrow$ & PRO-P $\uparrow$ \\
\midrule
\checkmark &  &  &  
& 94.2 & 98.2 & 97.7 & 61.1 & 88.6 \\
\checkmark & \checkmark &  &  
& 92.7 & 97.0 & 97.8 & 58.4 & 87.9 \\
\checkmark &  & \checkmark &  
& 94.8 & 98.2 & 97.8 & 62.5 & 89.0 \\
\checkmark &  &  & \checkmark 
& 97.2 & 99.0 & 97.7 & 65.0 & 89.9 \\
\midrule
\checkmark & \checkmark & \checkmark & \checkmark
& \textbf{97.4} & \textbf{99.1} & \textbf{98.0} & \textbf{65.3} & \textbf{90.2} \\
\bottomrule
\end{tabular}
}
\end{table}

\subsubsection{Component Analysis}
To evaluate the contribution of each architectural component, we conduct ablation experiments by removing individual modules from the full model while keeping all other settings unchanged. As shown in Table~\ref{tab:ablation_component}, removing any component consistently degrades performance, indicating that each module contributes to the effectiveness of the framework. Among the three components, the structure prior plays the most critical role, as its removal leads to the largest performance drop, suggesting that global structural guidance is essential for stabilizing cross-resolution feature alignment.

The detail prior mainly contributes to fine-grained anomaly localization. Removing it results in a moderate decrease in pixel-level metrics, indicating that local detail cues refine anomaly responses on top of the global structural guidance. In addition, the noise-aware augmentation improves robustness against nuisance patterns commonly observed in industrial environments; without it, the model becomes more susceptible to noise-like artifacts. The best performance is obtained when all components are jointly enabled, highlighting their cooperative roles in improving cross-resolution alignment and anomaly localization.

\begin{table}
\centering
\caption{Ablation study on individual components of HLGFA on MVTec AD. All models use Wide-ResNet50 as the backbone and the full loss objective.}
\label{tab:ablation_component}
\setlength{\tabcolsep}{6pt}
\resizebox{\textwidth}{!}{%
\begin{tabular}{l c c c c c}
\toprule
Variant
& AUROC-I $\uparrow$ & AP-I $\uparrow$
& AUROC-P $\uparrow$ & AP-P $\uparrow$ & PRO-P $\uparrow$ \\
\midrule
w/o Structure Prior & 86.6 & 94.7 & 90.1 & 45.4 & 67.7 \\
w/o Detail Prior & 95.5 & 98.4 & 97.7 & 63.4 & 89.6 \\
w/o Noise Aug & 96.7 & 98.7 & 97.7 & 57.0 & 88.2 \\
\midrule
Full Model & \textbf{97.4} & \textbf{99.1} & \textbf{98.0} & \textbf{65.3} & \textbf{90.2} \\
\bottomrule
\end{tabular}
}
\end{table}

\section{Conclusion}
\label{sec:conclusion}
In this paper, we proposed \textbf{HLGFA}, a novel unsupervised anomaly detection framework based on \emph{cross-resolution guided feature alignment}. Unlike reconstruction- or memory-based methods that explicitly model normal appearance, HLGFA exploits the inherent asymmetry between high- and low-resolution representations and converts their alignment inconsistency into reliable anomaly signals.

The key insight of HLGFA is that high-resolution features preserve stable structural semantics, whereas low-resolution features tend to over-generalize toward normal patterns. By introducing a structure–detail decoupled guidance mechanism, HLGFA refines low-resolution representations under high-resolution supervision, enhancing sensitivity to subtle anomalies while maintaining localization accuracy. Consequently, anomalies are revealed as regions where cross-resolution alignment breaks down, rather than being inferred from potentially unreliable reconstruction errors.

Extensive experiments on the MVTec AD and VisA benchmarks demonstrate the effectiveness of HLGFA. The proposed method achieves strong and consistent performance across datasets, particularly in pixel-level localization metrics such as AUROC and PRO. On MVTec AD, HLGFA attains competitive image-level detection performance while maintaining high-quality anomaly localization. These results highlight the robustness of the proposed cross-resolution alignment strategy for detecting diverse industrial defects.

From an industrial perspective, HLGFA is well suited for real-world AOI applications: it requires only normal samples for training, leverages pretrained backbones without task-specific fine-tuning, and produces stable, interpretable anomaly maps based on feature inconsistency. These properties make HLGFA especially practical for industrial scenarios characterized by diverse defects and scarce abnormal data.

In future work, we will explore extending cross-resolution alignment to multi-modal inputs and foundation models, as well as adaptive consistency modeling to further enhance robustness and anomaly confidence estimation.

\clearpage
%
%
\bibliographystyle{splncs04}
\bibliography{main}

\end{document}